\documentclass[a4paper, 10pt, conference]{ieeeconf}      

\IEEEoverridecommandlockouts                              

\usepackage[utf8]{inputenc}
\usepackage[T1]{fontenc}
\usepackage{microtype}

\usepackage{url}

\usepackage{todonotes}

\newcommand{\BUSSI}{\textsc{BUSSARD}}

\title{\LARGE \bf
  \BUSSI{} --\\
  Better Understanding Social Situations\\
  for Autonomous Robot Decision-Making
}

\author{Stefan Schiffer$^{1}$
  and
  Astrid Rosenthal-von der P{\"u}tten$^{1}$
  and
  Bastian Leibe$^{2}$%
  \thanks{$^{1}$Stefan Schiffer and Astrid Rosenthal-von der P{\"u}tten %
    are with the Chair Individual and Technology, %
    Department of Society, Technology, and Human Factors, %
    RWTH Aachen University, Aachen Germany
    {\tt\small \{stefan.schiffer, arvdp\}@itec.rwth-aachen.de}}%
  \thanks{$^{2}$Bastian Leibe %
    is with the Chair Computer Science 13 (Computer Vision), %
    Department of Computer Science, %
    RWTH Aachen University, Aachen, Germany
    {\tt\small leibe@vision.rwth-aachen.de}}%
}

\begin{document}

\maketitle
\thispagestyle{empty}
\pagestyle{empty}

\begin{abstract}

  We report on our effort to create a corpus dataset of different
  social context situations in an office setting for further
  disciplinary and interdisciplinary research in computer vision,
  psychology, and human-robot-interaction. %
  For social robots to be able to behave appropriately, they need to
  be aware of the social context they act in.  Consider, for example,
  a robot with the task to deliver a personal message to a person.  If
  the person is arguing with an office mate at the time of message
  delivery, it might be more appropriate to delay playing the message
  as to respect the recipient's privacy and not to interfere with the
  current situation.  This can only be done if the situation is
  classified correctly and in a second step if an appropriate behavior
  is chosen that fits the social situation. %
  Our work aims to enable robots accomplishing the task of classifying
  social situations by creating a dataset composed of semantically
  annotated video scenes of office situations from television soap
  operas.  The dataset can then serve as a basis for conducting
  research in both computer vision and human-robot interaction.

\end{abstract}

\section{Introduction}


Social robots are envisioned to interact with humans in social
scenarios, for instance as assistants helping with daily tasks,
reminding of appointments, and delivering messages.  To interact in a
socially competent way, a robot needs to consider the full social
context -- otherwise the acceptance of the social robot is at stake.



In this paper we report on first findings in our effort to create a dataset
of social situations that can be used to help robots understand such situations
better and that is useful for research in computer vision, psychology,
and human-robot interaction alike.

The rest of the paper is organized as follows.
First, we briefly review the state-of-the-art both in social robotics and in computer vision.
Then, we sketch our project and report on the current state and lessons learnt so far.
Finally, we give an outlook on what work lies ahead of us.

\section{Background and Related Work}

While context in human-robot interaction was initially merely a question of user location,
in recent years the term is increasingly understood as a more complex concept:
the system must draw conclusions from the perceived situation in order to understand the current context in its entirety.
This is crucial to develop the sophisticated context models that are necessary to realize a smart environment \cite{Robles-Kim_IJAST2010_147335}.
Just localizing users only provides one bit of information necessary to understand what is going on.
Two users in the same room can mean many things: they can be friends or foes, they could be of equal or different status, they could trust or distrust each other.
\cite{Riek-Robinson_SPM2011_5753098} describe the challenges and opportunities of socially intelligent agents,
and break down social context to the influencing factors situational context, social roles, social norms and cultural conventions.

Overall, %
a large number of variables influences which behavior is socially expected or desired in a particular situation, and which behavior is perceived inappropriate.
Human information processing manages to grasp this large amount of information in the shortest time and thus makes it possible to judge social situations at a glance.
Social cognition usually takes place automatically, unconsciously and with a minimum of cognitive effort \cite{Fritsche-EtAl_SozPsy2018}.
Hence, the judgment of a social situation comes naturally to humans, without considerable cognitive effort,
but they often cannot say explicitly why they decided one way or the other or which factors determined the decision \cite{Lugrin-EtAl_SocRob2019_10.1007/978-3-030-35888-4_52}.
%

Computer vision scholars identified the lack of a social perspective as one of the challenges in visual scene understanding:
``The scarcity of a more social, contextual perspective in the automated analysis of human-human interactions is also reflected
in computer vision literature, where interactions are typically reduced to visually and temporally well-defined events.''
\cite[p.~1]{Stergiou-Poppe_CVIU2019_2019102799}. 
%
Research in computer vision has already addressed some aspects of social scene understanding \cite{Stergiou-Poppe_CVIU2019_2019102799}. %
For instance, the presence of people in a scene can be detected and tracked automatically \cite{Voigtlaender-EtAl_CVPR2019_8953401,Pfeiffer-EtAl_GCPR2019_10.1007/978-3-030-33676-9_39},
their posture can be assessed \cite{Cao-EtAl_TPAMI2021_8765346,Sarandi-EtAl_TBBIS2021_9257071}, 
and hand gestures can be determined \cite{Pfister-EtAl_ACCV2014_10.1007/978-3-319-16865-4_35}. 
One important factor limiting research in this direction, which is also highlighted in recent surveys \cite{Stergiou-Poppe_CVIU2019_2019102799},
is the lack of large, well-curated and annotated datasets capturing complex social interactions.
They %
are too small
(e.g., only 20 scenes, dataset ``UT interaction'' by \cite{Ryoo-Aggarwal_2010_UT-Interaction-Data}), 
they are not embedded in real environments (e.g. only in front of static background, with fixed perspective, or scripted/staged interactions,
such as the ``SBU Kinetic Interaction'' datasets (by \cite{Yun-EtAl_2012_6239234}) and ``UT interaction''),
and/or they only assess a very limited set of events (e.g. only four classes of events like handshake, kiss, hug, high-five;
such as the datasets ``TV Human Interaction'' by \cite{PatronPerez-EtAl_2010_BMVC.24.50} 
and ``Hollywood2'' by \cite{Marszalek_CVPR2009_5206557}). 
\section{Approach and Lessons Learnt}

Within \BUSSI{}, we aim to learn social context models from video sequences of person-person interactions extracted from a video corpus of telenovela episodes.
The idea is to take advantage of repeated settings and known personal relationships between the different characters in order to provide rich annotations of the interaction context in a semi-automatic manner.
Automatic annotation of tv series episodes has already been explored by several research groups, starting from the work by \cite{Everingham_BMVC2009_2009545,PatronPerez-EtAl_2010_BMVC.24.50}. 
We proceed in a similar fashion, yet our focus is slightly different.

The general approach in \BUSSI{} then is on curating a dataset of annotated social situations.
On that data we first identify basic features (e.g. visually perceptible features
such as pose or gaze, temporal dynamics of events) as well as ontological information
(e.g., age, gender, relationship between people, social rules on how to behave in a given situation). %
Then we want to estimate which (combinations of these) features and ontological
information determines or allows for abstract judgements about the social situations
(e.g., whether there is a conflict or intimacy, a formal or informal gathering, etc.).
We want to explore both, how humans form such a judgement as well as how robots
can potentially classify social situations like humans do.

The basic features mentioned above will first be labeled manually or, if possible,
they will be automatically computed with computer vision and verified afterwards.
On top of the basic features we will record a "scene type" (e.g. business, leisure, ...)
and a "scene mood" (e.g. tense, joyful, intimate).
We are currently evaluating whether to use the adjectives from PANAS \cite{Watson-etAl_1988_PANAS} for the latter.
Also, we register a one sentence description for every scene (shown to clickworkers)
in order to see whether our vocabulary is appropriate or not.


So far, we have screened hundreds of hours of video data of more than
ten TV series. Only four series qualified to collect data from, and
even with these four series, only few scenes meet an extended set of
requirements on the video data. These requirements include sufficient
realism, no overacting, no zooming or moving camera effects and more.
This is  because the video material should be as close as possible
to what a robot will see in its future deployment scenarios.
We had to realize, that such requirements are rather exclusive
and only few scenes were usable.
As an example, we took the German TV Series "Sturm der Liebe".%
\footnote{\tiny\url{https://daserste.de/unterhaltung/soaps-telenovelas/sturm-der-liebe/index.html}} %
It is a telenovela and it is considered the most successful daily TV series in Europe.
From 100 episodes we could eventually only collect 6:34 minutes of video data.
Every clip has a length of 10 to fifteen seconds. Scene types range from
business settings to flirts and romantic encounters.

We are now in the process of labeling the scenes with background
knowledge and automatic and manual annotations for the basic features
before we give the scene to clickworkers to classify the scenes
according to higher-level properties such as ``intimate'' or ``tense''
but also to confirm the manually assigned basic features.

Once this data is being farmed, we plan to test both end-to-end
machine learning as well as a knowledge-based supervised learning
method to automatically classify scenes on a robot with these learnt
models and to see how they compare.

\addtolength{\textheight}{-12cm}   



%

\enlargethispage{19pt}
\section*{ACKNOWLEDGMENT}
{
  \small
We would like to thank Lena Plum, Luisa Escherich, Paula Winter, and
Archit Dharma for their support in screening, selecting, collecting,
and annotating the video data. %
This work presented in this paper was funded under grant number OPSF728
in the Open Seed Fund 2022 pogram of the Exploratory Research Space
at RWTH Aachen University.
}



\bibliographystyle{IEEEtrans}
\bibliography{BUSSARD}

\end{document}